\documentclass[lettersize,journal]{IEEEtran}
\usepackage{amsmath,amsfonts}
\usepackage{algorithmic}
\usepackage{algorithm}
\usepackage{array}
\usepackage[caption=false,font=normalsize,labelfont=sf,textfont=sf]{subfig}
\usepackage{textcomp}
\usepackage{stfloats}
\usepackage{url}
\usepackage{verbatim}
\usepackage{graphicx}
\usepackage{cite}
\usepackage{multirow}
\usepackage{amsmath}
\usepackage{picins}
\usepackage{booktabs}
\usepackage{url}
\usepackage{xcolor}
\usepackage{hyperref}
\usepackage{color, colortbl}
\definecolor{LightCyan}{rgb}{0.95,0.95,0.951}
\definecolor{LightCoral}{rgb}{1,0.94,0.91}
\definecolor{customcolor}{RGB}{0, 102, 204} 
\hypersetup{
    colorlinks = true,
    linkcolor={customcolor},
    citecolor = {customcolor},
    urlcolor = {black}
}
\usepackage{multicol, multirow}
\usepackage{nicematrix}
\begin{document}



\title{ContextDet: Temporal Action Detection with Adaptive Context Aggregation}

\author{ 
Ning Wang, \and 
Yun Xiao,
Xiaopeng Peng,
\and Xiaojun Chang, Senior Member, IEEE,\\
\and  Xuanhong Wang, \and and Dingyi Fang \thanks{
This work was partially supported by the National Science Foundation of China No. 62372371 and 61972315, and the Key Project of the Shaanxi Province International Science and Technology Cooperation Program No.2022KWZ-14. (Corresponding author: Yun Xiao) 

Ning Wang, Yun Xiao and Dingyi Fang are with the
School of Information Science and Technology, Northwest University. E-mails: nwang@stumail.nwu.edu.cn; yxiao@nwu.edu.cn; dyf@nwu.edu.cn

Xiaopeng Peng is with Rochester Institute of Technology, Rochester NY 14623, United States. Email: xxp4248@rit.edu

Xiaojun Chang is with the Australian Artificial Intelligence Institute,
University of Technology Sydney, Ultimo, NSW 2007, Australia Email:
cxj273@gmail.com

Xuanghong Wang is with the School of Communications and Information Engineering and School of Artificial Intelligence, Xi'an University of Posts and Telecommunications
China. Email:wxh@xupt.edu.cn
}
}

\maketitle

\begin{abstract}

Temporal action detection (TAD), which locates and recognizes action segments, remains a challenging task in video understanding due to variable segment lengths and ambiguous boundaries. Existing methods treat neighboring contexts of an action segment indiscriminately, leading to imprecise boundary predictions. We introduce a single-stage ContextDet framework, which makes use of large-kernel convolutions in TAD for the first time. Our model features a pyramid adaptive context aggragation (ACA) architecture, capturing long context and improving action discriminability. Each ACA level consists of two novel modules.  The context attention module (CAM) identifies salient contextual information, encourages context diversity, and preserves context integrity through a context gating block (CGB). The long context module (LCM) makes use of a mixture of large- and small-kernel convolutions to adaptively gather long-range context and fine-grained local features. Additionally, by varying the length of these large kernels across the ACA pyramid, our model provides lightweight yet effective context aggregation and action discrimination. We conducted extensive experiments and compared our model with a number of advanced TAD methods on six challenging TAD benchmarks: MultiThumos, Charades, FineAction, EPIC-Kitchens 100, Thumos14, and HACS, demonstrating superior accuracy at reduced inference speed. 

\end{abstract}


\section{Introduction}

\begin{figure}
  \centering
  \includegraphics[width=\linewidth]{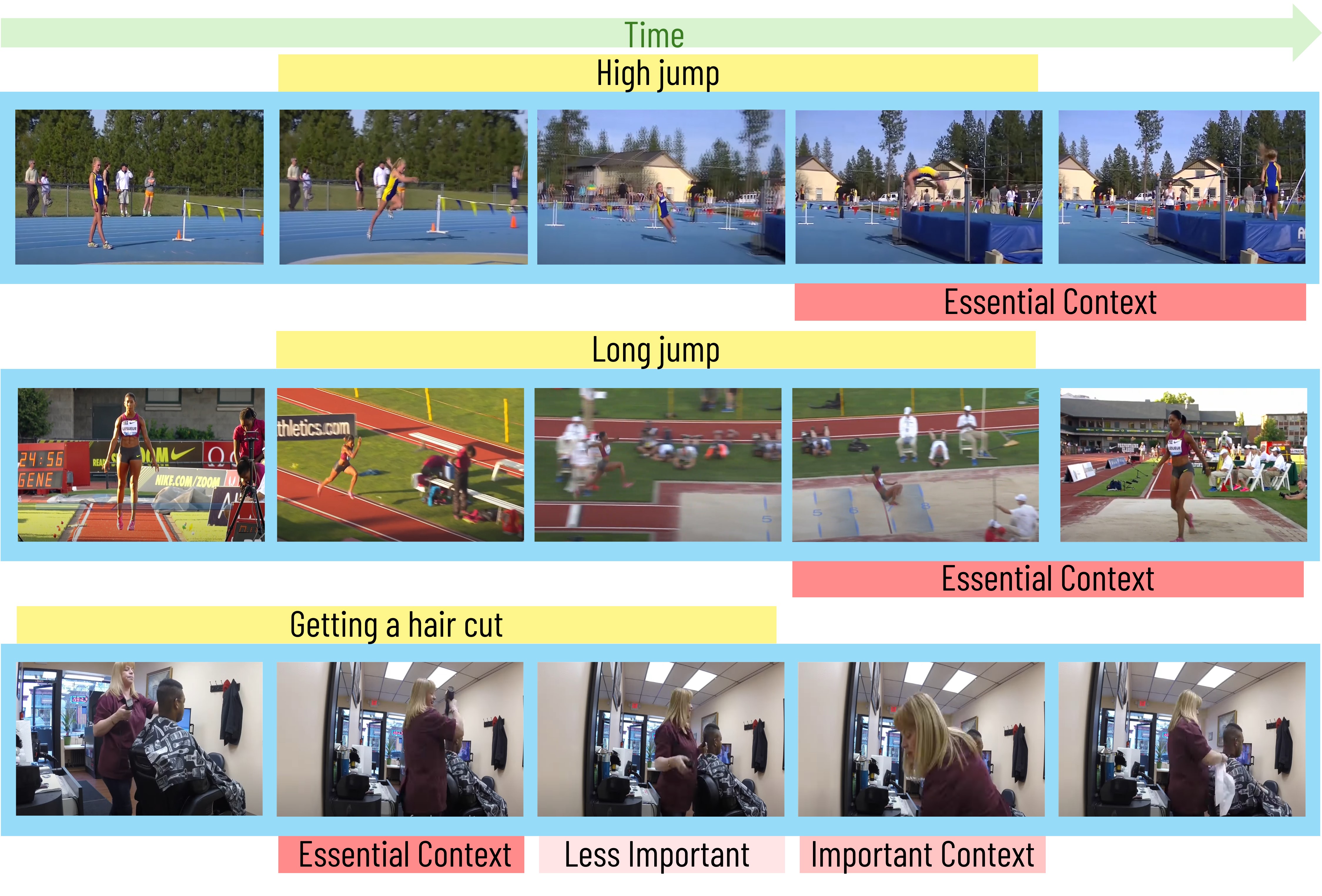}
  \caption{Accurate detection of action segments from a video sequence relies on discriminating salient information from long-term context. In additional to identifying the salient context in a long context, preserving context integrity and diversity as well as fine-grained local features are equally important. For example, distinguishing actions such as \textit{ high jump} and \textit{long jump} may benefit from recognizing the most salient and relevant contexts. On the other hand, ensuring the completeness of the long-range context which include a diverse relevance may also provide significant cues to improve the accuracy of the detection of actions like \textit{getting a hair cut}.}
  \label{fig1}
\end{figure}

\IEEEPARstart{T}{emporal} action detection (TAD) categorizes actions and identifies the boundaries of a video segment. TAD has been widely used in smart homes \cite{das2023explainable}, vision-language grounding \cite{soldan2021vlg}, video-based recommendation \cite{lee2017large}, multimedia retrieval \cite{zala2023hierarchical}, gaming technology \cite{gammulle2023continuous}, and more. Accurate localization of actions from videos remains challenging due to the varying lengths of action segments and the potentially ambiguous boundaries between them. 

The advancement of deep learning has significantly improved the performance of video action detection.  Contextual information, which are typically captured from frames that are adjacent to the action frame, provide relational information among frames. Capturing long-range temporal dependencies among video features can improve the performance of TAD for complicated actions. Transformers are favored in many natural language processing \cite{vaswani2017attention} and computer vision \cite{alexey2020image,arnab2021vivit,liu2021swin} applications  due to their superiority in capturing long-range dependencies through self-attention based token mixing. Transformer variants have been widely investigated for TAD tasks. ActionFormer \cite{zhang2022actionformer} is one of the representative methods that directly employs a multi-head Transformer for one-stage anchor-free TAD. Another ADSFormer \cite{li2024adaptive} makes use of a dual selective multi-head token mixer for channel selection and head selection in a pyramid structure to obtain important and discriminative features. Compare to convolutional neural network (CNN), however, several challanges remain presented in Transformer based TAD: 1) The rank loss. The self-attention provides a convex combination of the input features, which may increase the similarity and reduce the discriminability between the output features and thus affect the action detection accuracy negatively; 2) The quadratic computational cost of self-attention; and 3) The weaker inductive bias of Transformers requires orders of magnitude larger amount of training data \cite{touvron2021training}. 

To address the challenges that presented in Transformers, various CNN approaches have been studied for TAD. For example, the TriDet \cite{shi2023tridet} model substitutes self-attention blocks with convolution-based scalable granularity perception layers in a transformer-like architecture to improve the performance of action boundary discrimination. The TemporalMaxer \cite{tang2023temporalmaxer} method replaced the transformer encoder in ActionFormer \cite{zhang2022actionformer} with a combination of 1D convolutional and max-pooling. It minimizes feature redundancy and accelerate training speed while maximizing information from the extracted video clip features. Additionally, the detection of temporal saliency and aggregation of context are also explored in the weakly supervised temporal action detection by the use of max and average pooling, respectively \cite{zhao2023novel}.  Graph convolutional neural networks \cite{xu2020g} have also been employed to aggregate context by formulating video snippets and their relationships respectively as the node and edge of a graph. The edge of the graph is dynamically adjusted during training. Despite these advances, existing methods still face limitations in capturing contexts that satisfy optimal discriminability, rich details, and sufficient diversity at the same time. For example, the simple use of maxpooling in TempralMaxer limits its capability to discriminate richer and more diverse features other than the features with the highest value in the receptive fields. While the use of a two-branch CNN in Tridet may be helpful in improving the feature diversity, the feature discriminability may not be optimal. As shown in Fig. \ref{fig1}, capturing long-range and salient context information is crutial for accurate detection of complicated action segments. For example, distinguishing long jump and high jump actions is based on the most relevant contexts. However, relying only on the most relevant context may not be sufficient for the detection of some other actions. We illustrate such a scenario in the \textit{hair cutting} case, where the determination of the action requires contexts of a diverse range of saliency and relevance. 

In this work, we introduce a single-stage ContextDet model and demonstrate the first-time use of large-kernel convolutions in a Transformer-like architecture for temporal action detection. Our model consists of multiple levels of adaptive context aggregation (ACA) to extract multiscale pyramid features and is capable of capturing rich contextual information.  Each of the ACA levels consists of two novel modules: the context attention module (CAM) and the long context module (LCM). In the CAM typical self-attention was replaced by a two-branch design. The action features extracted by the K-branch are modulated by the context attention calculated by the Q-branch. In the Q-branch, a novel context gating block (CGB) was introduced to capture the salient contexts while preserving the context integrity and completeness. Although the use of 2D large kernel convolution has been explored for many computer vision tasks to replace the self-attention module, such as object detection \cite{ding2022scaling, cai2024poly, yu2024inceptionnext}, studies on the use of large kernel convolution for temporal action detection remain limited. In the LCM module,  we introduce the first-time design of 1D large-kernel convolution in TAD tasks to capture long-range contexts. Complementary small-kernel convolutions in LCM pay attention to fine-grained local features. Our proposed LCM consists of a mixture of large- and small-kernel convolution kernels. By varing the length of the large-kernel convolution, our model adaptively aggregates and modulates the neighboring context in an efficient manner. Through extensive experiments, we demonstrate that our ContextDet model outperforms alternative models in TAD. Specifically, our contributions are summarized as follows:



\begin{itemize}
    \item We propose a single-stage ContextDet model for temporal action localization, which discriminates the boundaries of action segments and predicts action categories from videos without the use of anchors or proposals.
\item The proposed ContextDet model has an adaptive context aggregation (ACA) pyramid architecture, where two novel modules are introduced at each level. The context attention module (CAM) features a context gating block (CGB), which dynamically selects the salient context while preserving the contextual completeness and diversity. The long context module (LCM) adaptively captures long-range contexts while paying attention to fine-grained local features through a mixture of large- and small-kernel convolutions.
    \item We demonstrate the use of 1D large-kernel convolution in temporal action detection for the first time. The varing lengths of the large-kernel convolution in the ACA feature pyramaid network allowing improved accuracy of at a reduced inference speed.
    \item The proposed ContextDet model outperforms a number of advanced TAD methods in qualitative and quantitative comparisons. State-of-the-art performance is achieved on six challenging benchmarks, they include MultiThumos \cite{yeung2015every}, Charades \cite{sigurdsson2016hollywood}, FineAction \cite{9934010}, EPIC-Kitchens 100 \cite{10.1007/s11263-021-01531-2}, Thumos14 \cite{Thumos14}, and HACSs \cite{zhao2019hacs}.
\end{itemize}


\section{RELATED WORK}
\label{relelated}
\subsection{Temporal Action Detection}
Temporal action detection identifies the start and end times stamps of video segments and predicts the categories of actions. Existing TAD methods include two-stage and single-stage approaches. The two-stage approach makes initial predictions based on a set of pre-generated proposals and refines the time stamps \cite{zhu2023contextloc++}. These methods focus on proposal generation. Anchor-based methods \cite{escorcia2016daps,buch2017sst, long2019gaussian}, for example, make use of densely distributed and multiscale anchors to generate proposals. Boundary-based methods \cite{lin2019bmn,lin2018bsn,liu2019multi,gong2020scale} predict the probability of each temporal point being either a start or an end of an action. In these algorithms, proposals are formulated and matched on the basis of the probabilistic scores. They are limited by the lack of end-to-end gradient flow \cite{liu2022end}. In contrast, single-stage methods do not require proposals, but detect action segments end-to-end. For example, TadTR \cite{liu2022end} and ReAct \cite{shi2022react} methods make use of a set of action queries to interact with the feature maps to detect action instances. Actionformer\cite{zhang2022actionformer} and Tridet\cite{shi2023tridet} take advantage of feature pyramid representations. Salient boundary features \cite{Lin_2021_CVPR1} are also explored to improve the performance of anchor-free methods. Here, we introduce a single-stage ContextDet model that adaptively aggregates salient context and sufficiently long contextual dependencies for more accurate temporal action detection at reduced inference speed.  



\begin{figure*}
  \centering
  \includegraphics[width=\linewidth]{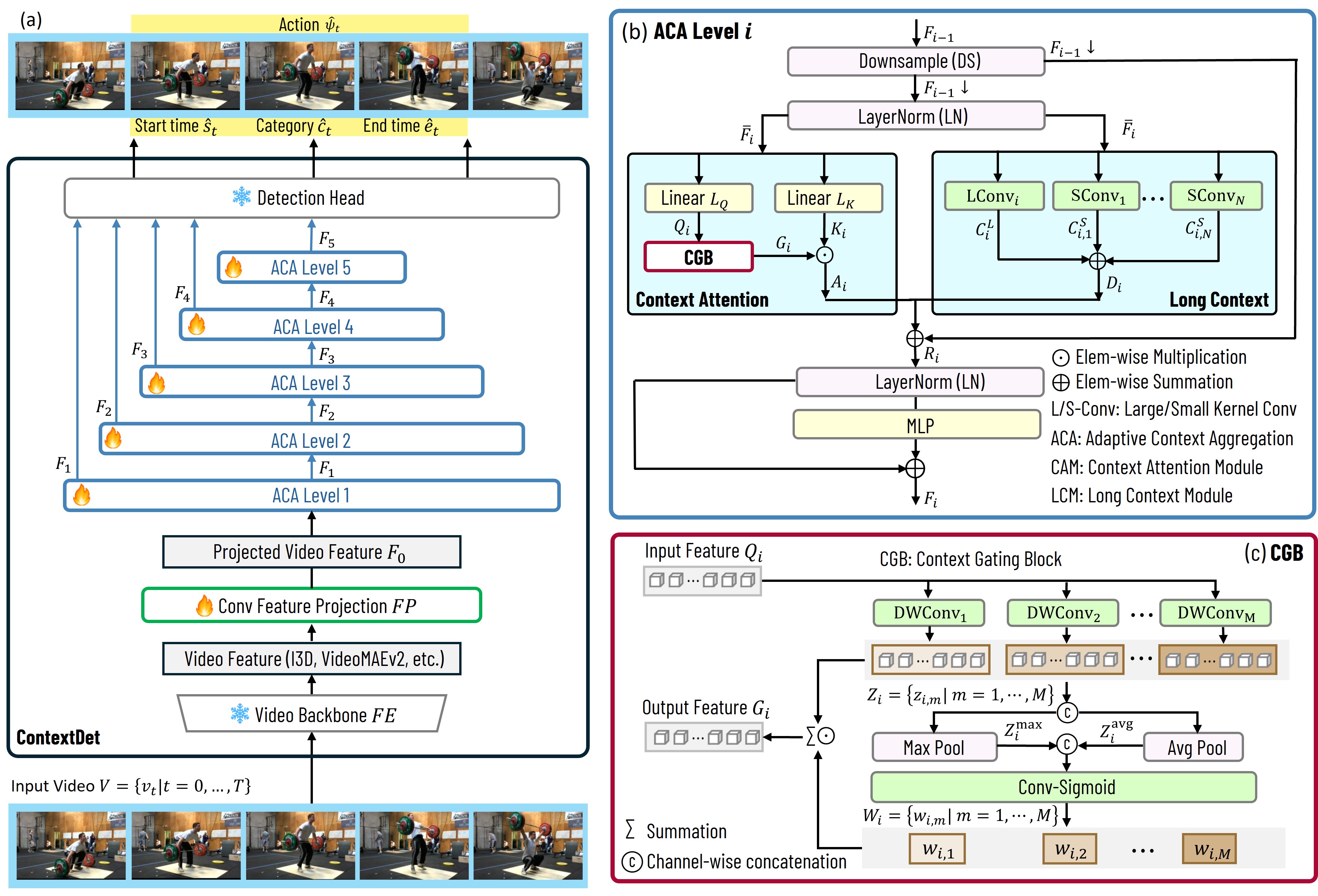}
  \caption{Illustration of the proposed single-stage ContextDet model for temporal action detection (TAD). (a) The architecture of the ContextDet model is comprised of a pre-trained video backbone  (e.g., I3D, VideoMAEv2, etc) as the feature extractor $FE$, a convolutional projection layer $FP$, five adaptive context aggregation (ACA) levels $i = 1,...,5$, and a pre-trained action detection head. The TAD pipeline starts from extracting video features from an input video $V = {v_t|t = 0,...,T}$ of a total number of $T$ frames. These video features are projected by the convolution layer $FP$ producing the projected video feature $F_0\in\mathbb{R}^{T_{0}\times D} $. This input feature passes the ACA layers, the outputs of each ACA layer are used to predict actions $\hat{\phi_t} = (\hat{s}_t, \hat{c}_t, \hat{e}_t)$. (b) Each ACA level starts with a downsampling layer, which reduces the dimension of the input feature $F_{i-1}$ by a half. The downsampled feature $F_{i-1}\downarrow$ the passes a layernorm (LN) layer. The following context attention module (CAM) consists of a Q-branch and a K-branch that diverted respectively by the linear layers $L_Q$ and $L_K$. The output of the Q-branch is sent into a context gating block (CGB), producing a gated feature $G_i$. The output of CAM $A_i$ is the result of the K-branch results $K_i$ modulated by the gated feature. The long context module (LCM) makes use of a large-kernel convolution and a number of $N$ small-kernal convolutions to capture the long-range context and fine-grained local features respectively. These context information are fused together producing the output $D_i$. The input video feature, the salient context and the long context information are then fused by a final LN and an MLP layer producing the output video feature $F_i$. (c) The illustration of the CGB, where $Z_i = \{z_{i,m}\}$ features are extracted by set of $M$ depth-wise convolution kernels $\mathrm{DWConv}_m$ and modulated by corresponding weights $W_i = \{w_{i,m}\}$ for $m = 1,...M$. There CNN kernels are varying in scales. The weights $W_i$ are calculated from the feature $Z_i$ by fusing the Max Pooling and the Average Pooling features via an additional Conv-Signoid layer. These weights modulate the CGB features to capture the context saliency that is most relevant to the action while preserving contextual integrity and diversity. }
  \label{figall}
\end{figure*}

\subsection{Large Kernel Neural Networks}
Many computer vision and multimedia tasks have benefited from the use of large window attention in Transformers as well as large convolution kernel in CNNs. The Swin Transformer\cite{liu2021swin}, for example, employs $7\times7$ to $12\times 12$ shifted window attention for object detection and classification. In the work of RepLKNet\cite{ding2022scaling} the size of the convolutional kernel was scaled to $31\times 31$ for object detection, where large-kernel CNNs demonstrated larger effective receptive fields than deep small-kernel models. In the recent PeLK net \cite{chen2024pelk}, an extremely large $101 \times 101$  peripheral convolution is introduced to increase the effective receptive field of CNNs at a significantly reduced number of parameters. The large selective kernel network (LSKNet) \cite{Li_2023_ICCV} decomposes dynamically large kernel convolutions into depth-wise convolutions to adjust its large spatial receptive field and model the context of various objects in remote sensing applications. The universal perception large kernel context network (UniRepLKNet) \cite{ding2023unireplknet} achieves improved performance in multiple modal applications, such as point clouds and audio. The parallel use of multiscale convolutional kernels has also been studied in computer vision tasks, such as object detection \cite{szegedy2016rethinking, liu2022convnet}. The inception networks \cite{cai2024poly, yu2024inceptionnext} splits features into several branches and applies the depth-wise convolution on each branch respectively. Although the practice of splitting the features reduces the computational cost and is effective for image classification tasks, we found that this technique tends to lower the TAD accuracies.  In this work, we introduce a long-context module (LCM) which makes the use of 1D large-kernel convolution for temporal action detection for the first time. The module consists of a mixture of large- and small- kernels, capturing long context and local feature variations at the same time. By reducing the size of the large kernels in the feature pyramid, we achieve improved accuracy at reduced inference speed.  


\subsection{Attention and Gating Mechanism}
 Machine learning and deep learning has been employed in diverse areas \cite{peng2024learning, peng2017randomized, peng2021cnn, peng2022computational, xiao2024multi, weng2009computer}, including the TAD tasks \cite{wang2023temporal}. The attention mechanisms, in particular, achieved remarkable success. In addition to the variances of vision Transformers where self-attention is employed, attentions have also been explored through the gating machenism in CNNs. For example, squeeze-and-excitation (SE) \cite{hu2018squeeze} and gather-excite (GE) \cite{hu2018gather} select salient features by squeezing spatial features into a channel descriptor and exited that descriptor.  The convolutional block attention module (CBAM) \cite{woo2018cbam} uses reweighed channels and spatial positions to adaptively modulate the feature map, achieving both channel and spatial attention. The local-relation net (LR-Net) \cite{hu2019local} adaptively determines feature aggregation weights based on the local pixel pairs.  Gated feature selections have also been investigated for capturing context information and action recognition \cite{zhou2024fease}.  For example, the CondConv\cite{yang2019condconv} and the dynamic convolution\cite{chen2020dynamic} methods utilize multiple parallel convolution kernels to adaptively extract features.  To capture the salient context while preserving its integrity, we present a context attention module (CAM) which fuse the CNN features with gated attention features at varying scales. Compared to channel grouping\cite{lin2023scale}, our model provides more accurate discrimination of features on different scales, allowing capturing diverse contextual information.  


\section{ContextDet Model}
\label{method}

\subsection{Model Architecture}
As illustrated in Fig. \ref{figall}(a), the proposed ContextDet model consists of four modules: a pre-trained video feature extraction backbone $FE$, a convolution projection layer $FP$, the multistage ACA module, and a pre-trained convolution-based detection head. The pre-trained video model (e.g., I3D \cite{i3d}, VideoMAEv2\cite{wang2023videomaev2}, etc.) extracts video features. Following that a projection layer embeds these features. The embedded features are then further fed into the multi-level ACA pyramid. Each ACA level is composed of a context attention module (CAM) and a long context module (LCM). In the CAM, we introduces a context gating block block to replace self-attention and capture the salient context. In the LCM, a mixture of large- and small-kernal convolution are employed to identify long context information without losing fine-grained local attention.  The multi-scale ACA features are passed to a pre-trained detection heads for action detection, which typically consists of a pair of decoupled classification and regression heads. 


\textbf{Feature Extraction and Projection} Given an untrimmed video having $T$ frames $V\in \mathbb{R}^{C\times H\times W\times T}$, each frame has a height $H$, width $W$, and number of channels $C$. The proposed ContextDet model detects a set of $U$ actions $\Psi =\left \{ \psi_u|u = 1,...,U \right \}$. Each action is denoted as $\psi_u = \left (s_u,e_u,c_u \right )$, where $s_u$ and $e_u$ are repectively the start and end point of the action ($s_u < e_u$), and $c_u$ is an action from a total number of $U$ action categories. Temporal features are extracted by the extraction backbone $FE$ and projected by the projection layer $FP$. The projection layer consists of two convolutional layers that are activated by the $Relu$ function. The projected input feature $F_0 \in\mathbb{R}^{T_0 \times D}$ is given by:
\begin{equation}
    F_{0} = FP(FE(V))
\end{equation}

\textbf{Adaptive Context Aggregation.} To capture a diverse range of relevant context for temporal action detection, we introduce in our ContextDet consists of five Adaptive Context Aggregation (ACA) stagets $i = 1,...,5$.  Each ACA level is composed of a downsampling (DS) layer, a context attention module (CAM), a long context module (LCM), an MLP layer, two LayerNorm (LN), and two skip connections. Denoting the input and output of each stage as $F_{i-1}$ and ${F_{i}}$ respectively, each of the ACA levels is given by:
\begin{equation}
\begin{split}
F_{i-1}\downarrow & = \text{DS}(F_{i-1})\\
\bar{F_{i}}  &=\text{LN}(F_{i-1}\downarrow)\\
R_{i} &= \text{CAM}(\bar{F_{i}}) + \text{LCM}(\bar{F_{i}}) + F_{i-1}\downarrow\\
F_{i} &= \text{MLP}(\text{LN}(R_{i})) + R_{i}
\end{split}
\end{equation}

\noindent Denoting $T_i$ and D are the number of temporal features and channel dimension respectively, at each stage  $F_{i-1}\in\mathbb{R}^{T_{i-1} \times D}$ is the input feature to the each ACA level and it is downsampled by a factor of two as $F_{i-1}\downarrow\in\mathbb{R}^{T_{i}\times D}$, where $T_i = T_{i-1}/2$

\textbf{Context Attention Module.} To extract the most relevant temporal context for action detection, a temporal context attention module (CAM) is introduced (see Fig. \ref{figall}(b)). In this module, context attention is calculated from the input video feature $A_i = \text{CAM}(\bar{F_{i}})$. Each CAM consists of two branches: a K-branch and a Q-branch. The K-branch extracts action features $K_i$ through a linear layer $L_K$:
\begin{equation}
   K_i=L_{k}\bar{F_{i}}
\end{equation}
\noindent In the Q-branch, the video feature passes through a linear layer $L_Q$ producing an output $Q_i$:
\begin{equation}
       Q_i = L_{Q}\bar{F_{i}}
\end{equation}

\noindent These Q-features are then fed into the context gating block (CGB) to calculate the gated attention $G_i = CGB(Q_i)$, the detail of which is described below. Each CGB block extracts multiscale features $z_{i,m}$ using a set of $M$ multiscale convolution kernels. These multiscale features are concatenated channel wise as:
\begin{equation}
\begin{split}
    z_{i,m}&=\text{GeLU}\big(\mathrm{DWConv}_{m}\left( Q_i \right)\big)\\
    Z_i &= \text{Concat}\big(\{z_{i,m}\}\big)
\end{split}
\end{equation}
where $m = 1,...,M$ and each of these depth-wise convolutional kernels has a distinct size. The max and average pooling are then applied to the concatenated feature to extract salient and average information. To capture rich and diverse contexts, we further concatenate the max and average temporal features:
\begin{equation}
\begin{split}
    Z_{i}^{\mathrm{max}}&=\mathrm{MaxPool}(Z_i)\\
    Z_{i}^{\mathrm{avg}}&=\mathrm{AvgPool}(Z_i)\\
    Z_{i}^{\mathrm{cat}} &= \text{Concat}\big([Z_{i}^{\mathrm{avg}};Z_{i}^{\mathrm{max}}]\big)
    \end{split}
\end{equation}
\noindent where the channel-wise max and average pooling are applied to the input features respectively. The max pooling captures salient contextual information, with the average pooling preserves feature integrity and completeness. The mixed feature $Z_{i}^{\mathrm{cat}}$ then passes the convolution-sigmoid layer to obtain the gating coefficients:
\begin{equation}
    W_i = \{w_{i,m}\} = \text{Sigmoid}\big(\text{Conv}(Z_{i}^{\mathrm{cat}})\big)
\end{equation}
\noindent The gated multi-scale temporal attention feature are given by:
\begin{equation}
    G_i=\sum_{m=1}^{M} z_{i,m}\odot w_{i,m}
\end{equation}
where $\odot $ represents the element-wise multiplication.
The output of CAM is given by K-features $K_i$ modulated the gated attention $G_i$ as:
\begin{equation}
\begin{split}
   A_{i}&= G_i\odot K_i
\end{split}
\end{equation}
\noindent where $G_i$ changes adaptively with respect to different inputs, and thus capturing context in a dynamic manner. 

\textbf{Long Context Module.} To capture long-range context without losing local details, we make use of a mixture of large- and small kenerl convolutions in a long context module (LCM) as shown in Fig. \ref{figall}(c). In order to capture the long context, we employ 1D large-kernel convolutions to expand the receptive field along the temporal direction. However, merely enlarging the convolution kernel leads to an only slight improvement in the detection performance of our model during the experiment process. While increasing the size of convolution kernel may increase the receptive field thus perception of longer context, an architecture with large-kernel convolution along may not be able to pay attention to fine-grained local features. To solve this issue, we introduce the parallel use of three smaller-kernel 1D convolutions as complementary to the large-kernel convolutions. Each of these small kernels has a length smaller than three. By fusing the results of a mixture of large- and small-kernel convolutions,  we are able to capture long-term context and fine-grained local feature at the same time. Each LCM at level $i$ is defined as $D_{i} = LCM(\bar{F}_{i})$, the details of which are written as:
\begin{equation}
\begin{split}
   D_{i} &= C^{L}_{i} + \sum_{n=1}^{N} C^S_{i, n}
\end{split}
\end{equation}
\noindent where the large- and small- convolution features are given respectively by:
\begin{equation}
\begin{split}
   C^L_{i}&=\text{GeLU}\big(\text{LConv}_{i}(\bar{F}_{i})\big)\\
   C^S_{i, n}&=\text{GeLU}\big(\text{SConv}_{n}(\bar{F}_{i})\big)
\end{split}
\end{equation}
\noindent where $\text{LConv}_{i}$ and $\{\text{SConv}_{n}| n = 1,..,N)\}$ are respectively a large-kernal convolution and a set of $N=3$ small-kernel convolutions at each layer. Here we use Gelu as activation functions for each convolution layer. Batch normalization has been employed with convolution \cite{ding2022scaling}. However, the use of batch normalization is observed to reduce the performance of our model, which might be explained by 1D convolutions having fewer parameters than 2D convolution. Additionally, we vary the size of the large convolution kernel at each ACA pyramid level to improve the diversity of receptive fields and efficiency. The size of the three small convolution kernels is kept fixed cross the pyramid. 


\textbf{Action Detection.}  Actions are decoded from a list of feature pyramid $ \left \{ F_i|i=1,2,...5\right\}$ by a detection head. Here we make use of a pre-trained detection head \cite{shi2023tridet} which consists of two disentangled heads respectively for classification and regression. The classification head predicts the probability $p(c_t)$ of an action $c_t$ at each time stamp $t$. The regression head predicts the duration of time $\Delta t_s$ and $\Delta t_e$ that lapses from the time stamp $t$ to the start point $\hat{s}_{t}$ and end point $\hat{e}_{t}$ respectively. The predicted video segment is written as:
\begin{equation}
\hat{\psi}_t=\left (\hat{s}_{t}, \hat{e}_t, \hat{c}_t\right )
\end{equation}
\noindent where
\begin{equation}
\begin{split}
    \hat{s}_{t}&=2^{i-1}\times \left (t-\Delta t_s  \right)\\
    \hat{e}_t &=2^{i-1}\times \left (t+\Delta t_e\right)\\
    \hat{c}_t&=\arg\max p(c_t)
\end{split}
\end{equation}

\begin{table*}[t]
  \centering
  \caption{Comparison of results on MultiThumos and Charades datasets.}
  \setlength{\tabcolsep}{10pt} 
\small
\begin{NiceTabular}{@{}>{\hspace{6.5pt}}l*{8}{|l|l|c|ccccc}@{}} 
    \toprule
  \multicolumn{1}{c|}{\multirow{2}{*}{Dataset}}&
  \multicolumn{1}{c|}{\multirow{2}{*}{Method}}&
  \multicolumn{1}{c|}{\multirow{2}{*}{Venue/Year}}&
  \multicolumn{1}{c|}{\multirow{2}{*}{Feature}}&
  \multicolumn{4}{c}{mAP @ tIoU (\%)}\\
  \cmidrule{5-8}
   &&&&0.2   & 0.5   & 0.7   &Avg\\
    \midrule
   &PDAN\cite{dai2021pdan} & WACV'2021& I3D (RGB)    & $-$ & $-$ & $-$ & 17.3     \\
   &MLAD\cite{tirupattur2021modeling} & CVPR'2021&  I3D (RGB)    & $-$ & $-$ & $-$ & 14.2     \\
  &MS-TCT\cite{dai2022mstct} &CVPR'2022  & I3D (RGB)    & $-$ & $-$ & $-$ & 16.2    \\
   &PointTAD\cite{tan2022pointtad}  &NeurIPS'2022  & I3D (RGB)   & 39.7 & 24.9 & 12.0 & 23.5    \\
   &ASL\cite{shao2023action} &ICCV'2023 & I3D (RGB)    & 42.4 & 27.8 & 13.7 & 25.5      \\
   MultiThumos&TemporalMaxer\cite{tang2023temporalmaxer} &Arxiv'2023 & I3D (RGB)    & 47.5 & 33.4 & 17.4 & 29.9      \\
   &TriDet\cite{shi2023tridet} &CVPR'2023& I3D (RGB)  & 55.7 & 41.0 & 23.5 & 36.2     \\
    
    &ADSFormer\cite{li2024adaptive}&TMM'2024 &I3D (RGB) &62.3&48.0&28.5&41.8  \\
       & \cellcolor{LightCyan}\textbf{ContextDet (ours)} &\cellcolor{LightCyan}2024 &\cellcolor{LightCyan}I3D (RGB)  &\cellcolor{LightCyan}63.0 &\cellcolor{LightCyan} 49.0 & \cellcolor{LightCyan}29.9 & \cellcolor{LightCyan}42.5   \\
   &TriDet\cite{shi2023temporal}&CVPR'2023 &VideoMAEv2& 57.7 & 42.7 & 24.3 & 37.5    \\
   &ADSFormer\cite{li2024adaptive}&TMM'2024 &VideoMAEv2 & \underline{64.4} &\underline{51.0} & \underline{31.7} &\underline{44.1}  \\
    & \cellcolor{LightCyan}\textbf{ContextDet (ours)}&\cellcolor{LightCyan}2024 &\cellcolor{LightCyan}VideoMAEv2 & \cellcolor{LightCyan}\textbf{65.6} & \cellcolor{LightCyan}\textbf{51.5} & \cellcolor{LightCyan}\textbf{31.8} & \cellcolor{LightCyan}\textbf{44.6}   \\
       \midrule
      & PDAN\cite{dai2021pdan} & WACV'2021& I3D (RGB)      & $-$ & $-$ & $-$ &8.5  \\
      &MS-TCT\cite{dai2022mstct} &CVPR'2022  &I3D (RGB)     & $-$ & $-$ & $-$ & 7.9\\
       Charades &PointTAD\cite{tan2022pointtad}  &NeurIPS'2022 & I3D (RGB)      &15.9     & 12.6     & 8.5     & 11.3 \\
        &ASL\cite{shao2023action} &ICCV'2023&I3D (RGB)        &24.5    & 16.5     &9.4     & 15.4 \\
         &TriDet\cite{shi2023tridet}&CVPR'2023 &I3D (RGB)     &\underline{27.1}  &\underline{20.4}  &\underline{13.2}  &\underline{18.4} \\
         &\cellcolor{LightCyan}\textbf{ContextDet (ours)}&\cellcolor{LightCyan}2024 &\cellcolor{LightCyan}I3D (RGB)    &\cellcolor{LightCyan}\textbf{30.3}    & \cellcolor{LightCyan}\textbf{22.9}     &\cellcolor{LightCyan}\textbf{14.0}    & \cellcolor{LightCyan}\textbf{20.3} \\
    \bottomrule
    \end{NiceTabular}
  \label{tab:MultiThumos and Charades}%
\end{table*}%

\begin{table}[t]
    \centering
    \caption{Comparison of results on FineAction dataset.}
\setlength{\tabcolsep}{7.8pt} 
\small 
\begin{NiceTabular}{@{}>{\hspace{6.5pt}}l*{5}{|c|cccc}@{}} 
    \toprule
     \multicolumn{1}{c|}{\multirow{2}{*}{Method}}&\multicolumn{1}{c|}{\multirow{2}{*}{Feature}}&\multicolumn{3}{c}{mAP @ tIoU (\%)}\\
     \cmidrule{3-5}
     && 0.5 & 0.75 &Avg \\
      \midrule 
     DBG\cite{DBG2020arXiv}& I3D&10.7	&6.4 &6.8\\
     G-TAD\cite{xu2020g}& I3D&13.7	&8.8&9.1\\
     BMN\cite{lin2019bmn}  &I3D	&14.4	&8.9	&9.3\\
     Actionformer\cite{wang2023videomaev2}&VideoMAEv2		&\underline{29.1}	&\underline{17.7}	&\underline{18.2}\\
    \multicolumn{1}{l|}{\cellcolor{LightCyan}\textbf{ContextDet (ours)}}&\cellcolor{LightCyan}VideoMAEv2&\cellcolor{LightCyan}\textbf{33.9}&\cellcolor{LightCyan}\textbf{20.5} &\cellcolor{LightCyan}\textbf{20.6}\\
\bottomrule 
    \end{NiceTabular}
    \label{tab:FineAction}
\end{table}

\section{Experiments}
\label{exce}
\subsection{Model Learning}
The model predicts the probability $p\left ( c_t \right ) $ for each action category, as well as the time lapses $\Delta t_s$ and $\Delta t_e$ from the current time $t$ to the action boundary. The loss function consists of a focal loss $\mathcal{L}_{cls}$   \cite{tian2022fully} for classification and an IoU loss $\mathcal{L}_{reg}$ \cite{rezatofighi2019generalized} for regression. The total $\mathcal{L}_{total}$ loss is given by:

\begin{equation}
\begin{split}
    \mathcal{L}_{\text{total}} &= \frac{1}{N_{\mathrm{pos}}} \sum_{t} \mathbb{I}_{c_t > 0} \cdot\left( \sigma_{IoU}\cdot\mathcal{L}_{\text{cls}} + \lambda\cdot \mathcal{L}_{\text{reg}} \right) \\
    &\quad + \frac{1}{N_{\mathrm{neg}}} \sum_{t} \mathbb{I}_{c_t = 0}\cdot \mathcal{L}_{\text{cls}}
\end{split}
\end{equation}
where $N_{\mathrm{pos}}$ and $N_{\mathrm{neg}}$ are the
number of positive and negative samples respectively. $\mathbb{I}_{c_t > 0}$ and $\mathbb{I}_{c_t = 0}$ denote respectively the time stamp of an action $c_t$ and its background. $\sigma_{IoU}$ is the temporal IoU between ground truth and predicted segment, and $\lambda$ is a coefficient that modulates the regression loss.

\subsection{Datasets}
We conducted evaluation of the proposed ContextDet model on six challenging datasets: MultiThumos \cite{yeung2015every}, Charades \cite{sigurdsson2016hollywood}, FineAction \cite{9934010}, EPIC-Kitchens 100 \cite{10.1007/s11263-021-01531-2}, Thumos14 \cite{Thumos14}, and HACS \cite{zhao2019hacs}. The MultiThumos and Charades are two densely multi-label TAD datasets, where the MultiThumos dataset includes 38,690 annotations for 65 types of sports action. The Charades dataset is a large-scale densely annotated multi-label dataset, including 9848 videos across 157 action categories. The FineAction is a fine-grained multi-label video dataset, which has 16,732 videos, 103,324 action instances, and 106 action categories. The EPIC-KITCHEN 100 dataset is a multi-label action dataset recorded in first-person view. It consists of 633 videos with a total number of 100 hours. It also involves a verb and a noun tasks, each having 97 and 300 categories respectively. The HACS and Thumos14 are two single-label datasets. The HACS is a large-scale action dataset, consisting of 49,485 videos and 122,304 daily life action instances. The Thumos14 dataset comprises 413 untrimmed videos, including 6,316 instances with 20 types of sport actions. 

\subsection{Evaluation Metrics}
Mean Average Precision (mAP) is a metric widely used to evaluate detection model performance. In our experiments, we use mAP at different tIoU thresholds in addition to the average-mAPs. The tIoU indicates the intersection over the union between ground truth and the predicted time intervals. The setting of tIoU follows the routines of the official guidelines and existing literatures \cite{zhang2022actionformer}\cite{shi2023tridet}\cite{wang2023videomaev2}.

\subsection{Training}
Experiments are conducted on an NVIDIA GeForce RTX 4090 GPU. Our model is trained with an AdamW \cite{girdhar2019video} optimizer on five datasets: MultiThumos, Charades, Thumos14,  EPIC-Kitchens 100 noun,  EPIC-Kitchens 100 verb. For each dataset, we train our model respectively for a total number of 46, 13, 43, 21, 19 epochs. It is found that warming up improves the convergence of our model, and we use 20, 5, 20, 5, 5 warm-up epochs for the corresponding dataset. The batch sizes are respectively 2, 16, 2, 2, 2, and the initial learning rate is set to $1e$-4.  For the FineAction and HACS dataset, we train our model for 16 and 10 epochs, including 7 warm-up epochs. The batch size in these two cases is 16 and the initial learning rate is $1e$-3. The learning rate is regulated by a cosine annealing scheduler \cite{Loshchilov2016SGDRSG} during the training. The number of layers in the pyramid is set to 6 for all datasets. The minimum length of the large kernel is kept at 5. The maximum lengths of the large kernels are capped respectively at 17 for Multithumos, Thumos14, and HACS datasets, 13 for the Charades and FineAction datasets, and 21 for the EPIC-Kitchens 100 dataset. In the post-processing stage, the SoftNMS \cite{bodla2017soft} method is used to discard inaccurate predictions.

\begin{table}[t]
    \centering
    \caption{Comparison of results on EPIC-Kitchens 100 dataset \\for verb and noun. }
\setlength{\tabcolsep}{7.3pt}
\small 
\begin{NiceTabular}{@{}>{\hspace{6.5pt}}l*{6}{|l|ccccc}@{}} 
    \toprule
    \multicolumn{1}{c|}{\multirow{2}{*}{Type}}
     &\multicolumn{1}{c|}{\multirow{2}{*}{Method}}
     &\multicolumn{4}{c}{mAP @ tIoU (\%)}\\
     \cmidrule{3-6}
     &&0.1&0.3&0.5&Avg\\
      \midrule 
      &BMN\cite{lin2019bmn}  &10.8 &8.4  &5.6 &8.4\\
    &G-TAD\cite{xu2020g} &12.1 &9.4  &6.5 &9.4\\
    \multicolumn{1}{c}{\multirow{4}{*}{Verb.}}&ActionFormer\cite{zhang2022actionformer} &26.6  &24.2 &19.1 &23.5\\
    &ASL\cite{shao2023action}&27.9&25.5&19.8&24.6\\
    &TemporalMaxer\cite{tang2023temporalmaxer}&27.8&25.3&19.9&24.5\\
    &DyFADet\cite{yang2024dyfadet} & 28.0 &25.6&\underline{20.8}&25.0\\
    &TriDet\cite{shi2023tridet} &\underline{28.6}  &\underline{26.1}  &\underline{20.8} &\underline{25.4}\\
    
    &\cellcolor{LightCyan}\textbf{ContextDet (ours)}&\cellcolor{LightCyan}\textbf{29.7}&\cellcolor{LightCyan}\textbf{27.2}&\cellcolor{LightCyan}\textbf{21.9}&\cellcolor{LightCyan}\textbf{26.6}\\
    
     \midrule 
     &BMN\cite{lin2019bmn}   &10.3  &6.2& 3.4 &6.5\\
    &G-TAD\cite{xu2020g} &11.0  &8.6  &5.4 &8.4\\
    \multirow{4}{*}{Noun.}&ActionFormer\cite{zhang2022actionformer} &25.2  &22.7  &17.0 &21.9\\
    &ASL\cite{shao2023action}&26.0 &23.4 &17.7& 22.6\\
     &TemporalMaxer\cite{tang2023temporalmaxer}&26.3&23.5&17.6&22.8\\
      &DyFADet\cite{yang2024dyfadet}&26.8 & 24.1 & \underline{18.5} & 23.4\\
    &TriDet\cite{shi2023tridet} &\underline{27.4}  &\underline{24.6}  &18.3 &\underline{23.8}\\
    
    &\cellcolor{LightCyan}\textbf{ContextDet (ours)}&\cellcolor{LightCyan}\textbf{27.6}&\cellcolor{LightCyan}\textbf{24.9}&\cellcolor{LightCyan}\textbf{19.1}&\cellcolor{LightCyan}\textbf{24.1}\\
\bottomrule 
    \end{NiceTabular}
    \label{tab:EPICKITCHEN 100}
\end{table}

\section{Results}
\textbf{MultiThumos and Charades. } We compare our ContextDet model with a number of advanced TAD methods on these two datasets in terms of detection mAPs. As shown in Table \ref{tab:MultiThumos and Charades}, our model achieves the highest accuracy at all tIoU thresholds. In particular, our model provides an average-mAP of 44.6\% and 42.5\% on MultiThumos for VideoMAEv2 \cite{wang2023videomaev2} and I3D \cite{i3d} features respectively, indicating an improvement of respective 7.1\% and 6.3\% in accuracies compared to the second-best TriDet \cite{shi2023tridet} model. We also achieve an average-mAP of 20.3\% on the Charades dataset using only the RGB features extracted by the I3D backbone, showing an improved accuracy of 1.9\% compared to the second-best TriDet. These two datasets feature a strong sequential correlation among the dense actions, which affirms our model's capability in adaptively aggregating contextual information for action 
understanding.


\begin{table}[t]
    \centering
    \caption{Comparison of results on  HACS dataset.}
\setlength{\tabcolsep}{7.3pt}
\small 
\begin{NiceTabular}{@{}>{\hspace{6.5pt}}l*{6}{|c|ccccc}@{}} 
        \toprule 
        \multicolumn{1}{c|}{\multirow{2}{*}{Method}}&\multicolumn{1}{c|}{\multirow{2}{*}{Feature}}&\multicolumn{4}{c}{mAP @ tIoU (\%)}\\
        \cmidrule{3-6}
     && 0.5 & 0.75 &0.95&Avg \\
        \midrule 
      SSN\cite{SSN2017ICCV}&I3D &28.8 &18.8 &5.3 &19.0\\
        LoFi\cite{NEURIPS2021_522a9ae9}&TSM   &37.8 &24.4 &7.3 &24.6\\
        G-TAD\cite{xu2020g} &I3D &41.1 &27.6 &8.3 &27.5\\
        TadTR\cite{liu2022end} &I3D   &47.1  &32.1  &10.9  &32.1\\
        BMN\cite{lin2019bmn} &SlowFast &52.5 &36.4 &10.4 &35.8\\
        TCANet\cite{qing2021temporal} &SlowFast  &54.1 &37.2 &11.3 &36.8\\
        TriDet\cite{shi2023tridet} &SlowFast   &56.7 &39.3 &11.7 &38.6\\
        TriDet\cite{shi2023tridet} &VideoMAEv2 &\underline{62.4} &\underline{44.1} &\underline{13.1} &\underline{43.1}\\

 \multicolumn{1}{l|}{\cellcolor{LightCyan}\textbf{ContextDet}} &\cellcolor{LightCyan}VideoMAEv2 &\cellcolor{LightCyan}\textbf{63.0} &\cellcolor{LightCyan}\textbf{44.7}&\cellcolor{LightCyan}\textbf{14.6}&\cellcolor{LightCyan}\textbf{43.8}\\
        \bottomrule 
    \end{NiceTabular}
    \label{tab:Hacs}
\end{table}

\textbf{FineAction. } For this dataset, VideoMAEv2 \cite{wang2023videomaev2} features are used and results are shown in Table \ref{tab:FineAction}, The FineAction dataset features fine-grained action of a rich diversity, which contains many overlapping actions (different fine-grained actions occur simultaneously). Despite its sensitive to contextual information, our model achieves an average accuracy of 20.6\%, exceeding  the second-best ActionFormer \cite{zhang2022actionformer} model by a significant 2.4\% in accuracy. This demonstrates that a carefully designed convolution network architecture can exceed the performance of Transformers in TAD.

\textbf{EPIC-Kitchens 100.} Experiments are performed on Slowfast\cite{feichtenhofer2019slowfast} features for this dataset. As shown in Table \ref{tab:EPICKITCHEN 100}, our model outperforms all other models in both the verb and noun subtasks. This demonstrates the robustness of our model in action detection in first-person view videos, which is typically degraded by background disturbances.

\textbf{HACS.} We make use of the VideoMAEv2\cite{wang2023videomaev2} features in the experiment on this dataset.  As shown in Table \ref{tab:Hacs}, we achieve an average-mAP 43.8\% , which exceeds the second-best Tridet \cite{shi2023tridet} model by 0.7\%. The HACS dataset contains a large number of long action segments. The improved performance of our model on this dataset reaffirms the strength of our model on capturing long-range temporal. The result also showcases the superiority of our model in detecting salient contextual information without losing its diversity and integrity, as well as the capturing of fine-grained local features.   

\begin{figure}[t]
  \centering
  \includegraphics[width=\linewidth]{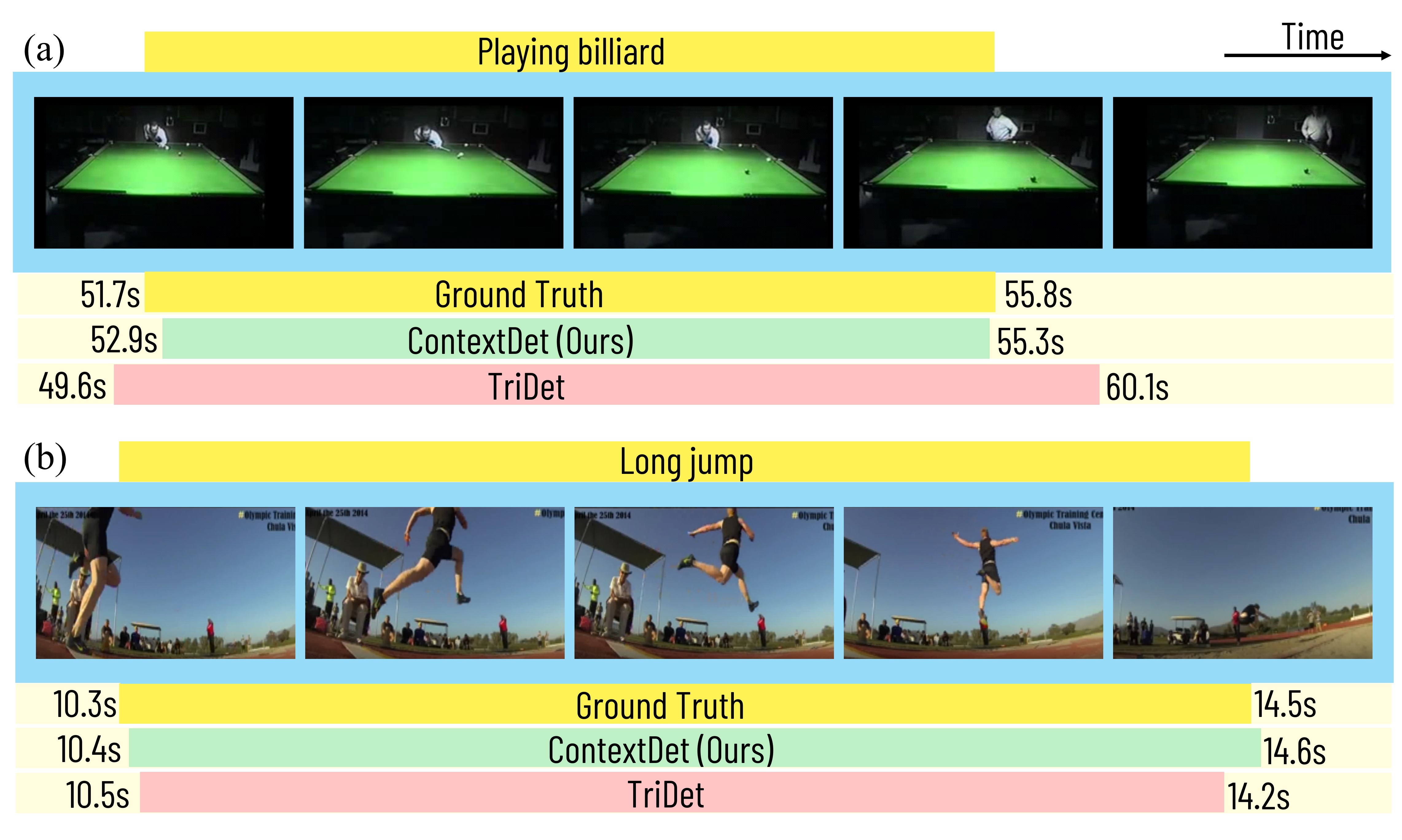}
  \caption{Qualitative evaluations of our ContextDet model and the Tridet \cite{shi2023tridet} model on two video clips from the THUMOS14 dataset, showcasing the actions \textit{playing billiard} and \textit{long jump} respectively. In each case, the yellow bar represents the ground truth, and the green and pink bars indicate respectively the detection results of our model and the Tridet model. Our model produces more accurate prediction of the starting point, the ending point, and the duration of the actions in both cases.}
  \label{example}
\end{figure}

\textbf{Thumos14.} The VideoMAEv2\cite{wang2023videomaev2} and I3D\cite{i3d} features are used in the experiment on this dataset. We showcase two qualitative evaluation of our ContextDet model in Fig. \ref{example}. Compared with the ground truth, both examples indicate that our model produces more accurate action prediction compared to the latest Tridet \cite{shi2023tridet} method. The qualitative results are presented in Table \ref{tab:Thumo14}. Our model achieves an average-mAP of 71.3\% with the use of VideoMAEv2 backbone, including a significant increase of 2\% in average-mAP and 1.7\% at tIoU=0.7 respectively compared with the Tridet model. With the use of I3D features, our ContextDet model outperforms all other models in terms of average-mAP. The Thumos14 dataset primarily consists of short sports clips, which affirms the efficacy of ContextDet in capturing short-segment temporal contextual information. 

\textbf{Latency.} We compared the number of model parameters and inference speed of our ContextDet model with two temporal action detection models: ActionFormer \cite{zhang2022actionformer} and Tridet \cite{shi2023tridet}. We report the the inference latency on THUMOS14 dataset using an input with the feature dimension 2304 × 2048. The inference time is averaged for 100 iterations and excluding another 20 iterations as GPU warmup times. As shown in Table \ref{tab:latency}, our model not only achieves the highest mAP but also has the fastest inference speed. Although our model has more parameters than Tridet, its computation method is more efficient, allowing for more effective use of computational resources. This accelerates the model's inference speed. Inference speed often plays a more crucial role in actual production.

\begin{table*}[htbp]
    \centering
    \caption{Comparison of results on Thumos14 dataset. }
  \setlength{\tabcolsep}{8.5pt} 
\small
\begin{NiceTabular}{@{}>{\hspace{6.3pt}}l*{8}{|l|l|c|cccccccc}@{}} 
        \toprule 
        \multicolumn{1}{c|}{\multirow{2}{*}{Type}}
        &\multicolumn{1}{c|}{\multirow{2}{*}{Method}}
        &\multicolumn{1}{c|}{\multirow{2}{*}{Venue/Year}}
        &\multicolumn{1}{c|}{\multirow{2}{*}{Feature}}
        &\multicolumn{6}{c}{mAP @ tIoU (\%)}\\
        \cmidrule{5-10}
     &&&&0.3 & 0.4 &0.5&0.6&0.7&Avg\\
        \midrule 
      &BMN\cite{lin2019bmn}&ICCV’2019&I3D  &56.0  &47.4  &38.8  &29.7  &20.5  &38.5\\
       &G-TAD\cite{xu2020g}&CVPR’2020&TSN&54.5&47.6&40.3&30.8&23.4&39.3\\
       &DBG\cite{DBG2020arXiv}&AAAI'2020&TSN&57.8&49.4&39.8&30.2&21.7&39.8\\
       &BC-GNN\cite{bai2020boundary}&ECCV'2020&TSN&57.1&49.1&40.4&31.2&23.1&40.2\\
       &A2Net\cite{A2Net} &TIP'2020&I3D&58.6&54.1&45.5&32.5&17.2&41.6\\
       &TCANet\cite{qing2021temporal}&CVPR’2021&TSN   &60.6  &53.2  &44.6  &36.8  &26.7  &44.3\\
      Two-Stage&BMN-CSA\cite{sridhar2021class}&ICCV'2021&TSP&64.4&58.0&49.2&38.2&27.8&47.7\\
      &RTD-Net\cite{tan2021relaxed}&ICCV’2021& I3D &68.3 &62.3 &51.9 &38.8 &23.7 &49.0\\
      &VSGN\cite{zhao2021video}&ICCV'2021&TSN &66.7 &60.4 &52.4 &41.0 &30.4 &50.2\\
      &MUSES\cite{Liu_2021_CVPR}&CVPR'2021&I3D&68.9&64.0&56.9&46.3&31.0&53.4\\
      &Disentangle\cite{ldclr}&AAAI'2022&I3D&72.1&65.9&57.0&44.2&28.5&53.5\\
      &SAC\cite{yang2022structuredattentioncompositiontemporal}&TIP'2022&I3D&69.3&64.8&57.6&47.0&31.5&54.0\\
      &ContextLoc++\cite{zhu2023contextloc++}&TPAMI'2023	&I3D	&74.4	&68.2	&58.7	&46.3	&30.8 & 55.7\\
        &TC-TAD\cite{xia2023exploring}&TMM'2023&I3D&81.6&78.4&71.4&60.0&45.1&67.5\\
         \midrule 
       &AFSD\cite{Lin_2021_CVPR1}&CVPR'2021&I3D    &67.3 &62.4 &55.5 &43.7 &31.1 &52.0\\    &TAGS\cite{nag2022proposalfreetemporalactiondetection}&ECCV'2022&I3D&68.6&63.8&57.0&46.3&31.8&52.8\\
        &ReAct\cite{shi2022react} &ECCV’2022&TSN &69.2 &65.0 &57.1 &47.8 &35.6 &55.0\\
        &TadTR\cite{liu2022end}&TIP‘2022&I3D &74.8 & 69.1 &60.1 &46.6 &32.8&56.7 \\			
        &Self-DETR\cite{kim2023self}&ICCV'2023&I3D &74.6 &69.5 &60.0 &47.6 &31.8 &56.7\\
        & TALLFormer\cite{cheng2022tallformer}& ECCV’2022& Swin &76.0 &- &63.2 &- &34.5 &59.2\\
        &Actionformer\cite{zhang2022actionformer}&ECCV’2022&I3D&82.1 &77.8 &71.0 &59.4 &43.9 &66.8\\
       &TransGMC\cite{yang2023gated}&TMM'2023&I3D& 82.3& 78.8& 71.4& 60.0& 45.1&67.5\\
       &ASL\cite{shao2023action}&ICCV'2023&I3D&83.1	&79.0	&71.7	&59.7	&45.8 &67.9\\
       One-Stage&DyFADet\cite{yang2024dyfadet}&ECCV'2024&I3D&84.0&80.1&72.7 &61.1&47.9&69.2\\
       &Tridet\cite{shi2023tridet}&CVPR'2023&I3D&83.6& 80.1 &72.9 &62.4 &47.4& 69.3\\
       &MFAM-TAL \cite{tip2024}&TIP'2024&I3D&83.0&79.5&73.8&62.5&48.2&69.4\\
       &ADSFormer\cite{li2024adaptive}&TMM'2024&I3D&82.9&79.9&73.4&62.8&47.8&69.4\\
       &\cellcolor{LightCyan}\textbf{ContextDet (ours)}&\cellcolor{LightCyan}2024&\cellcolor{LightCyan}I3D&\cellcolor{LightCyan}83.9&\cellcolor{LightCyan}80.0&\cellcolor{LightCyan}73.2&\cellcolor{LightCyan}62.1&\cellcolor{LightCyan}48.2&\cellcolor{LightCyan}69.5\\
       &Actionformer\cite{wang2023videomaev2}&ECCV'2022&VideoMAEv2& 84.0 &79.6 &73.0 &63.5 &47.7 &69.6\\
       &MFAM-TAL \cite{tip2024}&TIP'2024&VideoMAE&84.6&80.8&73.5&61.7&48.6&69.8\\
       &Tridet\cite{shi2023temporal}&CVPR'2023&VideoMAEv2 &84.8	&80.0	&73.3	&63.8	&48.8&70.1\\
       &DyFADet\cite{yang2024dyfadet}&ECCV'2024 &videoMAEv2& 84.3 &-&73.7&-&\underline{50.2}&70.5\\
       &ADSFormer\cite{li2024adaptive}&TMM'2024&videoMAEv2&\underline{85.3}&\underline{80.8}&\underline{73.9}&\underline{64.0}&49.8&\underline{70.8}\\       
& \cellcolor{LightCyan}\textbf{ContextDet (ours)} &\cellcolor{LightCyan}2024&\cellcolor{LightCyan}VideoMAEv2 &\cellcolor{LightCyan}\textbf{85.6}&\cellcolor{LightCyan}\textbf{81.2}&\cellcolor{LightCyan}\textbf{74.4}&\cellcolor{LightCyan}\textbf{64.5}&\cellcolor{LightCyan}\textbf{50.5} &\cellcolor{LightCyan}\textbf{71.3}\\
        \bottomrule 
    \end{NiceTabular}
    \label{tab:Thumo14}
\end{table*}

\begin{table}[t]
    \centering
    \caption{Comparison of computation cost vs. accuracy on THUMOS14.}
\setlength{\tabcolsep}{6.5pt}
\small 
\begin{NiceTabular}{@{}>{\hspace{6.4pt}}l*{6}{|c|c|cccc}@{}} 
        \toprule 
        \multicolumn{1}{c|}{\multirow{2}{*}{Method}}&{\multirow{1}{*}{Params\vspace{-1em}}}&\multicolumn{1}{c|}{\multirow{1}{*}{Latency\vspace{-1em}}}&\multicolumn{3}{c}{mAP @ tIoU (\%)}\\
        \cmidrule{4-6}
     &(MB)& (ms) &0.5 & 0.7&Avg\\
        \midrule 
     ActionFormer\cite{SSN2017ICCV} &29.2 &84.9 &71.0 &43.9 &66.8 \\
        Tridet\cite{shi2023tridet}  &\textbf{16.0} &75.1&72.9 &47.4&69.3\\
        \multicolumn{1}{l|}{\cellcolor{LightCyan}\textbf{ContextDet}}  &\cellcolor{LightCyan}19.7 &\cellcolor{LightCyan}\textbf{65.7 }&\cellcolor{LightCyan}\textbf{73.2}&\cellcolor{LightCyan}\textbf{48.2}  &\cellcolor{LightCyan}\textbf{69.5}\\
        \bottomrule 
    \end{NiceTabular}
    \label{tab:latency}
\end{table}

\section{Ablation Study}
To evaluate the architecture design and learning strategies of the proposed ContextDet model, three ablation studies are conducted on the Thumos14 dataset \cite{Thumos14}.

\textbf{Ablation on Model Architecture.} To validate the effectiveness of LCM and CAM modules, we performed ablation on these two modules. We include a baseline model \cite{Lin_2021_CVPR1} (Method 1) into the comparison, which uses the same detection head. Additionally, we replaced LCM and CAM modules with the convolution-based scalable granularity perception (SGP) layer (Method 2) in TriDet \cite{shi2023tridet}. The dimensions of intermediate features, the number of layers in the pyramid feature layer, and the length of each layer remain the same for all models. The ablation results are provided in Table \ref{tab:Ad1}. In cases where the LCM and the CAM module are used individually (Methods 3 and 4), our model shows a significantly improved average-mAP by 4.5\% and 4.9\%, respectively. Furthermore, using either our LCM or CAM module outperforms the SGP by 0.3\% and 0.7\%, respectively. The combined use of our LCM and CAM modules (Method 5) provide an even higher accuracy, with the average-mAP outperforming the baseline and SGP by 5.4\% and 1.2\% respectively.

\textbf{Ablation on CGB Kernel Sizes.} To determine the number and size of convolution kernels that bring out the best performance of our model, several combinations of the number and lengths of the kernels are examined in context gating block (CGB) of the context attention module (CAM). The results of five different combinations are presented in Table \ref{tab:Ad2}. It is found that the optimal performance is achieved with the use of three kernels (1, 3, 5).  


\textbf{Ablation on LCM Kernel Sizes.} We also study the impact of large- and small- kernels in LCM to the TAD accuracies in Table \ref{tab:Ad3}. The first two rows showcase the use of large kernels alone, where the large kernel at each five ACA level has a length of 5 and 17 respectively. Although increasing the size of larger kernels leads to an improved accuracy by 0.4\%, the average-mAP in both cases is lower than the use of CAM alone (see Table \ref{tab:Ad2}). This might be explained by the loss of local details without any small kernels. This observation is affirmed by a significant improvement in accuracy through the combined of the same large kernels with small kernels. Compared with the second row, adding a set of three small kernels (1, 1, 3) to the same set of large kernels results in an increase in accuracy by 0.4\% in the third row. An even higher accuracy is achieved by varying the sizes of large kernels across the ACA pyramid (see the fourth row). Compared to the fixed large kernel sizes, the varying large kernel sizes may align better with the varying feature sizes in the pyramid. 




\begin{table}[t]
    \centering
    \caption{Ablation studies on CAM and LCM modules of our ContextDet model, baseline \cite{Lin_2021_CVPR1}, and SGP layer \cite{shi2023tridet} on Thumos14.}
\setlength{\tabcolsep}{6.3pt}
\small 
\begin{NiceTabular}{@{}>{\hspace{6.0pt}}c*{8}{|c|c|c|ccccc}@{}} 
    \toprule
    \multicolumn{1}{c|}{\multirow{2}{*}{Method}}&
    \multicolumn{1}{c|}{\multirow{2}{*}{SGP}}&
    \multicolumn{1}{c|}{\multirow{2}{*}{CAM}}&
    \multicolumn{1}{c|}{\multirow{2}{*}{LCM}}&
    \multicolumn{4}{c}{mAP @ tIoU (\%)}\\
    \cmidrule{5-8}
     &&&&0.3&0.5&0.7&Avg\\
      \midrule 
     1&&&&81.4 &69&43.5&65.9\\
     2&\checkmark&&&84.8 &73.3&48.8&70.1\\
    3&&\checkmark&&84.6&73.8&49.4&70.4\\
    4&&&\checkmark&85.6&73.8&49.6&70.8\\
\multicolumn{1}{c|}{\cellcolor{LightCyan}5}&\cellcolor{LightCyan}&\cellcolor{LightCyan}\checkmark&\cellcolor{LightCyan}\checkmark&\cellcolor{LightCyan}\textbf{85.6}&\cellcolor{LightCyan}\textbf{74.4}&\cellcolor{LightCyan}\textbf{50.5}&\cellcolor{LightCyan}\textbf{71.3}\\
\bottomrule 
\end{NiceTabular}
\label{tab:Ad1}
\end{table}

\begin{table}[t]
    \centering
    \caption{Ablation studies of the CGB kernel sizes of our ContextDet model on Thumos14.}
\setlength{\tabcolsep}{12pt}
\small 
\begin{NiceTabular}{@{}>{\hspace{18pt}}c*{5}{|cccccc}@{}} 
    \toprule
    \multicolumn{1}{c|}{\multirow{2}{*}{CGB Kernel Sizes}}&
    \multicolumn{4}{c}{mAP @ tIoU (\%)}\\
    \cmidrule{2-5}
     &0.3&0.5&0.7&Avg\\
      \midrule 
     (1,3)&84.7 &73.3&49.2&70.1\\
     (3,5)&85.2 &73.1&48.3&70.2\\
\multicolumn{1}{c|}{\cellcolor{LightCyan}(1,3,5)}&85.6&\cellcolor{LightCyan}\textbf{74.4}&\cellcolor{LightCyan}\textbf{50.5}&\cellcolor{LightCyan}\textbf{71.3}\\
    (3,5,7)&85.6&74.9&49.5&71.1\\
     (1,3,5,7)&\cellcolor{LightCyan}\textbf{85.7}&74.1&49.1&70.8\\
\bottomrule 
    \end{NiceTabular}
    \label{tab:Ad2}
\end{table}

 \begin{table}[t]
    \centering
    \caption{Ablation studies of the LCM kernels of our ContextDet model on Thumos14.}
\setlength{\tabcolsep}{7.0pt}
\small 
\begin{NiceTabular}{@{}>{\hspace{6.5pt}}l*{5}{|cccccc}@{}} 
    \toprule
    \multicolumn{1}{c|}{\multirow{2}{*}{LCM Kernel Sizes}}&
    \multicolumn{4}{c}{mAP @ tIoU (\%)}\\
    \cmidrule{2-5}
     &0.3&0.5&0.7&Avg\\
      \midrule 
     (5,5,5,5,5) w/o SConv &84.8 &73.0&48.0&69.8\\
     (17,17,17,17,17)  w/o SConv  &84.7 &74.0&49.0&70.2\\
    (17,17,17,17,17) w. SConv &85.0&74.2&49.5&70.6\\
    \multicolumn{1}{c|}{\cellcolor{LightCyan}(1,3,5) w. SConv}  &\cellcolor{LightCyan}\textbf{85.6}&\cellcolor{LightCyan}\textbf{74.4}&\cellcolor{LightCyan}\textbf{50.5}&\cellcolor{LightCyan}\textbf{71.3}\\
\bottomrule 
    \end{NiceTabular}
    \label{tab:Ad3}
\end{table}


\section{Error Analysis}
\begin{figure*}[!t]
\centering
\captionsetup[subfloat]{font=tiny}
\subfloat[Video Clip 1]{\includegraphics[width=9cm]{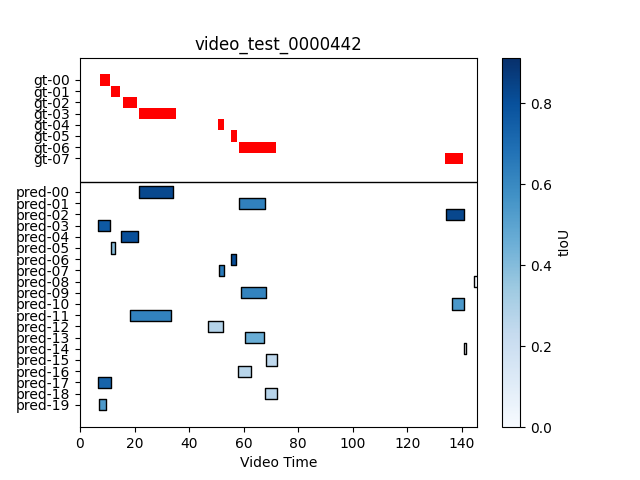}%
\label{Condconv}}
\hfil
\subfloat[Video Clip 2]{\includegraphics[width=9cm]{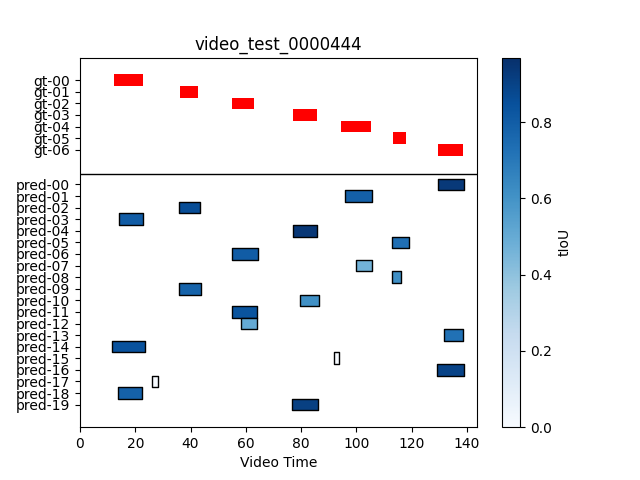}%
\label{Mixture}}
\hfil
\subfloat[Video Clip 3]{\includegraphics[width=9cm]{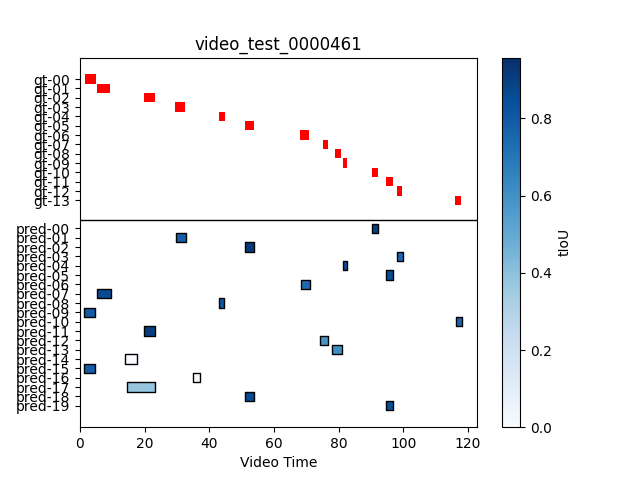}%
\label{Filter1}}
\hfil
\subfloat[Video Clip 4]{\includegraphics[width=9cm]{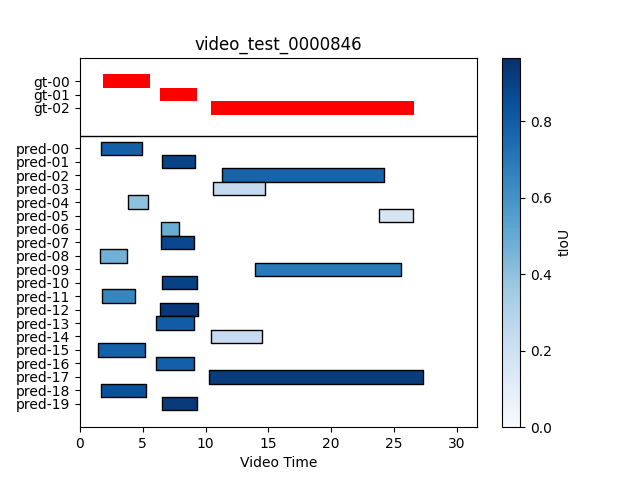}%
\label{Filter2}}
\hfil
  \caption{Qualitative results of our ContextDet model with VideoMAEv2 \cite{wang2023videomaev2} features on four video clips (a)-(d) from the Thumos14 test set. The red bars above the line represent the ground truth, and the blue bars below showcase the predicted action segments with the top 20 accuracies. The darkness of the color indicates the degree of overlapping of the results with the ground truth. }
  \label{visual}
\end{figure*}

A video clip is typically characterized by its length, coverage, and the number of instances. The length indicates the duration of a video in seconds. The coverage represents the length of an action instance normalized by the length of an entire video. The number of instances indicates the total number of actions of the same category in a video. Qualitative results of four Thumos14 video clips are shown in Fig. \ref{visual}(a)-(d) for twenty predicted segments of the top scores. The scores are determined by the maximum tIoU between the real and predicted actions. Compared to the ground truth (red), our CondextDet model (blue) produces boundary detection with minimal discrepancies. The quantitative diagnostic analysis \cite{alwassel2018diagnosing} of sensitivity, false positives, and false negatives are provided for Thumos14 video clips, each having a different length. Here we divide the video into five sets according to its coverage and length, respectively: extra short (XS), short (S), medium (M), long (L), and extra long (XL). We also divide the videos into four intervals based on the number of instances as: extra small (XS), small (S), medium (M), and large (L).

\begin{figure*}[t]
  \centering
  \includegraphics[width=0.97\linewidth]{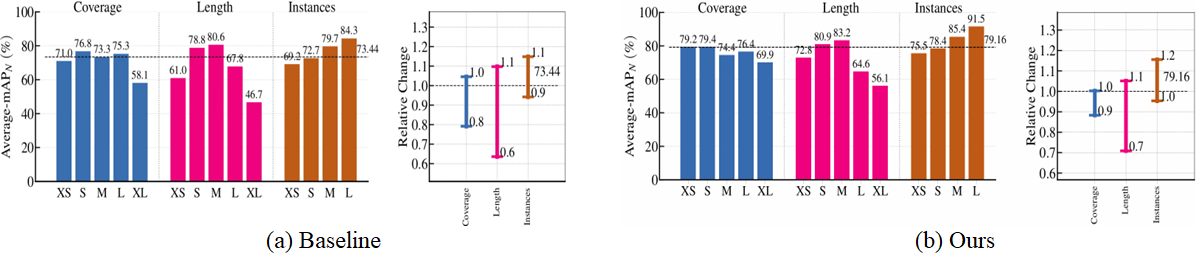}
  \caption{The sensitivity analysis of (a) the baseline model \cite{Lin_2021_CVPR1} and (b) our ContextDet model to action characteristics.  Left: each bar measures the average-$\mathrm{mAP_N}$ value at tIoU=0.5 on a subset of Thumos14 dataset that features a particular action characteristic. The dotted lines indicate the mean average-$\mathrm{mAP_N}$. Right: A summary of the left, where the sensitivity is given by the difference between the max and min average-$\mathrm{mAP_N}$ values.}
  \label{sensitive}
\end{figure*}

\textbf{Sensitivity.} As shown in Fig. \ref{sensitive}, our method outperforms the baseline \cite{Lin_2021_CVPR1} by 5.7\% in terms of the average $\mathrm{mAP_N}$ of the coverage, length, and instance measures (see the dotted lines in Fig. \ref{sensitive}(a) and (b)). Compared to the baseline model, our model also shows reduced relative sensitivity changes across the three metrics, which confirms the robustness of our model.

\begin{figure*}[t]
  \centering
  \includegraphics[width=0.97\linewidth]{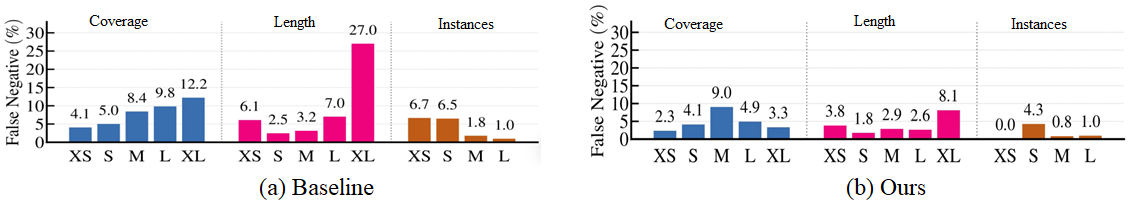}
  \caption{The false negative (FN) analysis of the (a) the baseline model \cite{Lin_2021_CVPR1} and (b) our ContextDet model, including  the probability of model omissions (false negatives) under three different metrics: coverage, length, and instance volume of a video. Significant reduction of these errors are observed using our model.}
  \label{FN}
\end{figure*}

\textbf{False Negative.} The false negative (FN) profiles are shown in Fig. \ref{FN}, which provides an indication of misdetected samples. Compared to the baseline \cite{Lin_2021_CVPR1} model shown in Fig. \ref{FN}(a), our model shown in Fig. \ref{FN}(b) reduces false negatives by a large margin in almost all cases. For the \textbf{coverage}, Our model provides reductions in the mean FN rate by 3.2\%. While our model exhibits a slightly higher FN rate in the M coverage, our FN rates are 1.8\%  and 8.9\% lower than the baseline for the XS and XL coverage. For the \textbf{length}, our FN rate is 5.3\% lower than the baseline in mean, and 2.3\% and 18.9\% lower for the XS and XL lengths. For the number of \textbf{instances}, our FN rates are respectively 2.5\% lower than the baseline in overall mean value, with the FN rate of our model reaches almost zero for the single action (XS) video. These improvements may be attributed to the advancement of our model in capturing long-range context information without compromising its integrity and diversity, as well as local features. 

\begin{figure*}[t]
  \centering
  \includegraphics[width=0.97\linewidth]{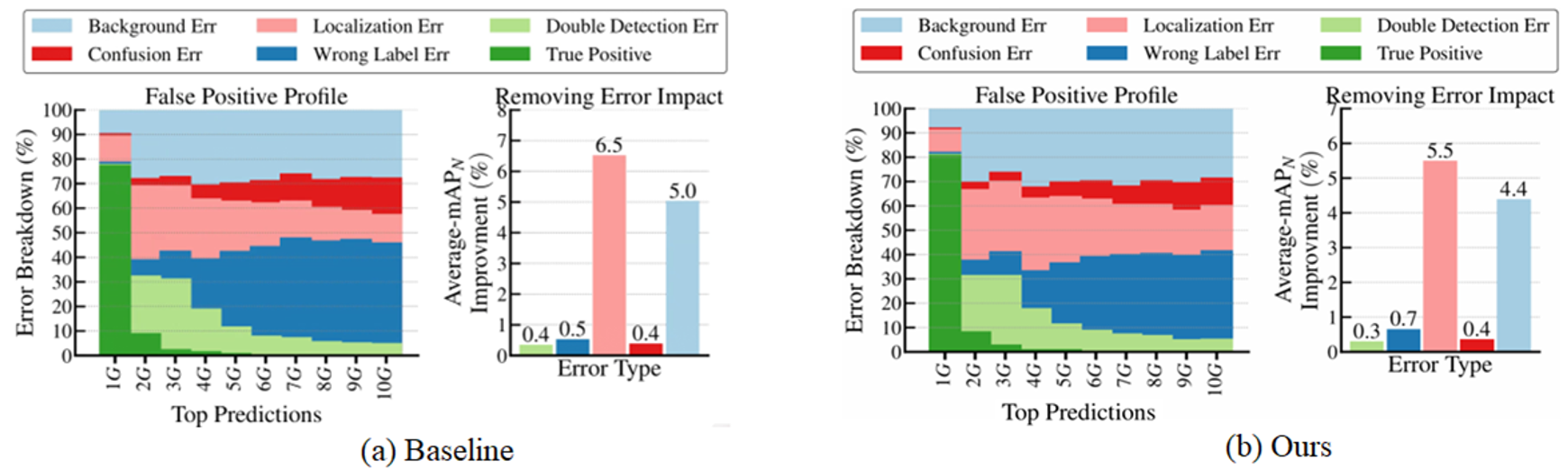}
  \caption{The false positive (FP) analysis of (a) the baseline model \cite{Lin_2021_CVPR1} and (b) our ContextDet model. Left: The FP profiles, each demonstrates the FP error breakdown in the top-10G predictions; Right: The improvements of average $\mathrm{mAP_N}$ from removing predictions that caused by different type of error. The localization errors (pink bar) and background errors (light blue bar) have the most impact.}
  \label{FP}
\end{figure*}

\textbf{False Positive.} The average $\mathrm{mAP_N}$ values relies on the predictions rankings. We show in each left figure of Fig. \ref{FP} the false positive (FP) profiles as functions of top-G predictions at tIoU=0.5, where $G$ is the number of ground truth. We divide the top-10G predictions into ten equal bins and showcase the breakdown of the five FP error types in each. While the true positive rate takes up the majority in both the baseline \cite{Lin_2021_CVPR1} and our model for 1$G$ predictions, our model outperforms the baseline with an FP value exceeds 80\%. Compared with the baseline, the wrong label error of our model is also notably lower across all top-G predictions, indicating the strength of our model in detection accuracy and robustness against action categories. Each right figure of Fig. \ref{FP}(a) and (b) showcase the average-mAP in cases where the predictions that cause one of the five types of errors are removed respectively. Compared to the baseline model, our model provides improvements in average-mAP by 5.5\% and 4.4\%  respectively for cases where the localization and background errors are removed.



\section{Conclusion}
\label{con}
In this work, we introduced a single-stage ContextDet model for temporal action detection based on a dynamically gated pyramid convolution neural network. Our model makes use of large-kernel convolutions in TAD for the first time to increase receptive field and capture long context. Through the combined use of max and average pooling, a mixture of large- and small kernels, as well as varying large kernel sizes across the pyramid, our model also provides an adaptive context aggregation to ensure the context integrity, context diversity, and fine-grained local features. We evaluated our model on six challenging datasets: MultiThumos, Charades, FineAction, EPIC-Kitchens 100, Thumos14, and HACS. Our model outperformed a number of advanced TAD algorithms in extensive experiments and ablation studies, and state-of-the-art accuracy and efficiency are demonstrated. The performance of our model may benefit from more advanced video feature extraction backbone and detection heads to reduce the localization and background errors. Future work may also include an end-to-end training of our model with these modules. 


\bibliographystyle{IEEEtran}
\bibliography{ref}

\end{document}